# Streamlining Advanced Taxi Assignment Strategies based on Legal Analysis

Holger Billhardt, José-Antonio Santos, Alberto Fernández, Mar Moreno,
Sascha Ossowski, José A. Rodríguez

CETINIA, Universidad Rey Juan Carlos, Madrid, Spain
`{holger.billhardt, joseantonio.santos, alberto.fernandez,`
`mar.rebato, sascha.ossowski, joseantonio.rodriguez}@urjc.es`

**Abstract.** In recent years many novel applications have appeared that promote the provision of services and activities in a collaborative manner. The key idea behind such systems is to take advantage of idle or underused capacities of existing resources, in order to provide improved services that assist people in their daily tasks, with additional functionality, enhanced efficiency, and/or reduced cost. Particularly in the domain of urban transportation, many researchers have put forward novel ideas, which are then implemented and evaluated through prototypes that usually draw upon AI methods and tools. However, such proposals also bring up multiple non-technical issues that need to be identified and addressed adequately if such systems are ever meant to be applied to the real world. While, in practice, legal and ethical aspects related to such AI-based systems are seldomly considered in the beginning of the research and development process, we argue that they not only restrict design decisions, but can also help guiding them.

In this manuscript, we set out from a prototype of a taxi coordination service that mediates between individual (and autonomous) taxis and potential customers. After representing key aspects of its operation in a semi-structured manner, we analyse its viability from the viewpoint of current legal restrictions and constraints, so as to identify additional non-functional requirements as well as options to address them. Then, we go one step ahead, and actually modify the existing prototype to incorporate the previously identified recommendations. Performing experiments with this improved system helps us identify the most adequate option among several legally admissible alternatives.

## 1 Introduction

With the advent of new Information and Communication Technologies (ICT), over the last decade new economic models have emerged that are based on sharing or exchanging underused assets and services among customers. Terms like "sharing economy",

"collaborative economy", or "collaborative consumption" [1] refer to such models. Even though many challenges for systems and applications in this field can be adequately addressed from a technical point of view, they often clash with a legal regulatory framework that was developed decades ago, thus hindering their smooth integration into our societies.

Systems in this field usually try to instil a more efficient use of available resources by promoting the shared use of free capacities. This is usually done by dynamically forging contracts and enacting transactions among stakeholders through agreement processes that are mediated by a distributed ICT platform. From the perspective of the Agreement Technologies (AT) paradigm, such systems are conceived as large-scale open distributed systems, where users delegate (part of) their decision-making to intelligent components called *software agents*, in order to be able to cope with the high frequency of interaction [2]. The ICT platform is conceived not only to *enable* communication among users and/or agents, but also to *regulate* interactions, thereby giving shape to the space of agreements that can effectively be reached. Autonomy, interaction, mobility, and openness are key characteristics that need to be dealt with in a robust and timely manner. Such ICT platforms can be implemented as an intelligent computing infrastructure that relies on some sort of institutions and institutional agents in order to avoid or discourage undesirable actions, or on an information system that strategically provides recommendations in an environment with a significant level of uncertainty [3,4]. On these grounds, several applications have been built in domains such as logistics, transportation, or emergency management [5,6,7].

Work usually focuses on showing the technical feasibility of such systems, and on analysing their performance (not only from the perspective of the platform owner, but also from the point of view of the users and their software agents). However, regulatory challenges that allow for their smooth integration into our societies are often overlooked. As [8] outlines, especially novel applications based on AT have a wide range of legal question and problems, which must be addressed if such systems are ever meant to be applied to the real world. In [9], we argued that the implications of such a legal analysis are twofold: on the one hand, it identifies as to how far clarifications or changes of legal regulations would be required; on the other hand, it provides further insights into modifications needed in the smart platform itself.

In this manuscript, we present a case study that shows how systematic legal analysis of a collaborative application can reveal both additional legal and technical requirements and how the latter, in turn, can be used to extend its functionality and to foster its applicability in a real-world setting. For this purpose, we set out from a system that we developed previously in this field, in particular, a taxi coordination service that mediates between individual (and autonomous) taxis and potential customers [10].

The contribution of the present work is twofold. Firstly, we analyse the viability of the aforementioned system from the viewpoint of current legal restrictions and constraints. From the analysis, we derive aspects that would require changes in terms of legal regulation, and point to some technical issues that can or should be changed in the initial proposal. Secondly, we go one step ahead, and actually modify our previous system to incorporate the previously identified recommendations. Performing experiments with this improved system helps us identify the most adequate option among several legally admissible alternatives.

The remainder of the manuscript is organised as follows. Section 2 briefly summarises related work, focusing especially on existing approaches related to an advanced and potentially collaborative management of taxi fleets. In section 3, after briefly summarising the taxi assignment approach presented in [10], we apply the steps proposed in [9] to analyse this approach, and obtain a high-level description of system operation suitable for analysis by legal experts. In section 4, we examine the different operational principals with regard to their legal compliance, in order to flesh out potential problems and identify options to overcome them, both from a technical and from a regulatory perspective. Upon this background, section 5 outlines various prospective modifications to the initially proposed taxi assignment approach and presents experimental results to analyse their performance. Finally, section 6 concludes this article and points to future lines of work.

## 2    Related Work

Traditionally, customer requests are performed by phone calls to a taxi control centre, which then dispatches a taxi through radio communication. Often a *first-come first-served (FCFS)* strategy is applied to assign taxis to customers, i.e. customers are assigned to available taxis in the order of their appearance. Efforts to develop improved heuristic dispatching rules in fully controlled environments are quite old [11]. Nowadays, in the taxi dispatching domain, the *nearest-taxi/nearest-request (NTNR)* strategy is often used as baseline. It assigns customers to the closest available taxis if there are more available taxis than customers (like *FCFS*), but allocates each available taxi to the closest waiting customer if there are less taxis than customers [12, 13].

The availability of real-time GPS data streams as well as support for Vehicle-to-Vehicle (V2V) and Vehicle-to-Infrastructure (V2I) communication has spawned interest in novel ICT platforms that support taxi fleet management. Many works put forward assignment strategies that aim at reducing the customers' waiting times. Initial work by Lee, Wang, Cheu, and Teo [14] revealed the advantages of considering real-time information (on taxis and traffic conditions) to assign taxis to customers based on estimated travel time rather than distance. Maciejewski, Bischoff and Nagel [15] propose to calculate the optimal assignment among idle taxis and pending requests at certain intervals or whenever new events (new customer/available taxi) occur. Zhu and Prabhakar [16] analyse how suboptimal individual decisions lead to global inefficiencies and propose an assignment model based on network flow. Zhan, Qian and Ukkusuri [17] use a graph-based dispatching technique in a system-wide recommendation system for taxi services, whose performance is then evaluated with real-world large-scale taxi trip data. Vazifeh et al [18] put forward a strategy based on minimum cost maximum bipartite matching and evaluate it with a dataset of 150 million taxi trips taken in the city of New York over one year. Their approach is particularly interesting as it was implemented as a simple urban app (without ride sharing) and did not require changes to regulations, business models, or human attitudes towards mobility to become effective.

Other approaches are driven by goals different from average customer waiting times. BAMOTR [19] provides a mechanism for fair assignment of drivers, where fair assignment is intended to minimize the differences in income among the taxi drivers. For that purpose, the authors minimize a combination of taxi income and extra waiting time. Gao et al. [20] propose an optimal multi-taxi dispatching method with a utility function that combines the total net profits of taxis and waiting time of passengers. They also consider different classes of taxis. Meghjani and Marczuk [21] propose a hybrid path search for fast, efficient, and reliable assignment to minimize the total travel cost with a limited knowledge of the network.

Other lines of work focus on taxi demand prediction with the goal of helping taxis to quickly find closer passengers or of balancing supply and demand of taxis in an area of interest (e.g. [22, 23]). The approach by Miao et al. [24, 25] is particularly interesting in relation to our approach, as they treat the problem of dispatching vacant taxis towards current and future demands while minimizing total idle mileage. Their approach is based on forecasting the uncertain spatial-temporal taxi demands in a region.

Another vibrant line of research refers to taxi ridesharing (e.g., [26, 27, 28]), even though the challenges of such a scenario are slightly different from the problem addressed in this manuscript.

Our work from [10], that the present manuscript sets out from, also tries to find assignments from taxis to pending customers that globally minimize the expected waiting times of customers. However, it differs from the aforementioned approaches as it considers the possibility of modifying an existing assignment when a taxi has been dispatched but has not yet picked up the corresponding customer. A similar approach has shown good performance for managing emergency medical where assignments between ambulances and patients need to be adapted on-the-fly [29]. One of the few other works that follows a similar path is [30], where reassignment is possible during some time interval based on negotiation among vehicle agents, but always before a pick-up order is sent to the taxi and the customer is informed.

As AI-based solutions are becoming increasingly popular, there is a growing awareness that smart applications need to be conceived, and perhaps even designed, with a multidisciplinary perspective in mind [31]. Research into legal and ethical implications of smart applications and (semi-)autonomous systems is of foremost interest in recent days [32].

Such a perspective has already been taken on by works in the field of Agreement Technologies (AT). In AT applications, human users delegate (part of) their decision making to intelligent components called *software agents*, in order to be able to cope with the high frequency of interaction. These software agents, in turn, make use of a sandbox of technologies, including automatic semantic alignment, norms and institutions, negotiation and argumentation, as well as trust and reputation, so as to (semi-)autonomously take decisions and forge agreements on behalf of their human users [2]. Casanovas [8] analysed the implications of such systems from the perspective of relational justice. In legal requirements engineering research [33] the focus is usually on specifying languages and methods to specify legal regulations, in order to identify their implications on a new system. However, in practice, legal implications are often considered only once a protype has already been built. As outlined in [9], in such cases a semi-formal specification of system behaviour is often sufficient to perform a legal analysis that examines compatibility with a legal framework and reveals options for adjusting a prototype accordingly.

The contribution of the present manuscript can thus be conceived from two perspectives. On the one hand, it constitutes a case study of how legal analysis can be used to identify options for streamlining a prototype of a smart taxi assignment system on the basis of legal requirements. On the other hand, it shows how this streamlining can actually be achieved by extending and further structuring existing methods in the prototype, and evaluating their adequacy and efficiency through simulation experiments.

## 3    Reassignment of customers among taxis

Urban mobility is one of the main concerns in big cities, and one of the main actors involved in the daily traffic activity in urban areas are taxi fleets. They consist of several thousands of vehicles in big cities (e.g. about 15,000 taxis in Madrid, Spain).

Two of the main goals of a taxi fleet are (i) to reduce the response time (e.g., the time between a customer call and the moment a taxi arrives at the customer's location) and (ii) to reduce costs of empty movements (e.g., movements taxis have to make in order to get to the location of customers). The provision of efficient methods for taxi assignment to customers is a challenge that can contribute to reducing distances of empty trips with the resulting decrease of traffic flow, pollution, and time.

As mentioned above, typically used assignment strategies like *FCFS* or *NTNR*, but also more sophisticated strategies proposed in the literature, are based on fixed assignments, i.e., once a taxi accepts a passenger, the dispatching is irreversible.

In the following section we will summarize the taxi assignment proposal from [10] that overcomes this restriction and allows for the reassignment of dispatched but not yet occupied taxis.

### 3.1    Assignment and compensation

The idea underlining the approach proposed in [10] exploits the fact that, at certain moments, a reassignment of customers among taxis may be beneficial from a global (cost) perspective. Fig. 1 depicts an example of such a situation. The figure shows two customers (C1 and C2) and two taxis (T1 and T2) in a geographic area. The destinations of the customers are called $Dest_{C1}$ and $Dest_{C2}$, respectively.

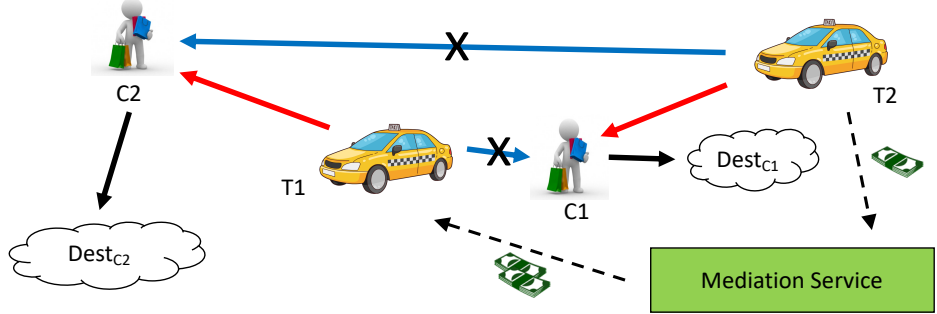

**Fig. 1.** Example of cost reduction through reassignment

Now let's suppose that taxi T1 is assigned to C1 and T2 to C2 (shown by the solid lines). In this case, the overall cost for serving both customers is determined by the driving distances dist(T1,C1) + dist(C1,Dest$_{C1}$) + dist(T2,C2) + dist(C2,Dest$_{C2}$). As shown schematically in the figure, this overall cost is higher than in the alternative assignment (doted lines): dist(T1,C2) + dist(C2,Dest$_{C2}$) + dist(T2,C1) + dist(C1,Dest$_{C1}$). In particular, the difference is due to the different distances from the taxis to the customers. Thus, a reassignment of C1 from T1 to T2 and of C2 from T2 to T1 would improve (reduce) the overall cost.

Furthermore, let's suppose a typical pricing scheme for taxi services where a customer only pays a fare for the distance/time she is moving in the taxi, and maybe some additional fixed price for entering the taxi. That is, the customers do not pay for the distance the taxis have to travel to reach their initial locations. Instead, the cost of this movement is afforded by the taxi drivers. In this case, an agreement between both taxis to simply exchange their customers would probably not be reached, because one taxi (in this case T1) would object, since the distance (and thus, the cost) for attending the new customer (C2) would increase (we assume for the moment that destinations of customers are unknown). In order to convince T1 to accept the exchange of customers, we would need to compensate her (at least) for the additional cost and the extra time spent. On the other hand, T2 would benefit from the exchange, both, in service time and cost. Thus, in comparison to attending customer C2, T2 would agree to give away some of its benefit. In particular, T2 should be willing to pay the travel cost it saves by attending C1 instead of C2 (then still T2 would earn in service time).

The situation described in the example is just one of many other possible situations where the overall service cost may be reduced by exchanging assigned customers among two or more taxis. In many such situations the compensations needed in order to convince taxi drivers who will be worse off after the exchange can be balanced by the amount of money other drivers would be willing to pay for the exchange.

To summarize the compensation scheme proposed by Billhardt et al. in [10], we use the following notations. Let $t$ be a taxi and $k$ a customer. Let $d_{tk}$ denote the distance or time units for taxi $t$ to reach the location of customer $k$, $d_{kD}$ the distance (time) of the trip from the location of customer $k$ to her destination, and $d_{tkD} = d_{tk} + d_{kD}$ the distance (time) of the entire trip. Furthermore, assume a pricing scheme as described previously, where *cost* represents the operational cost of moving the car per distance or time unit (including fuel, and other costs), *fcost* is a fixed cost the customer pays for any taxi trip, and *fare* is the fare per distance or time unit a customer pays for the trip from her origin to her destination. Then, the revenue of a trip for taxi $t$ attending customer $k$ is calculated as follows:

$$Revenue(t,k) = fcost + fare \cdot d_{kD} - cost \cdot d_{tkD} \tag{1}$$

Assuming some compensation $c$ when taxi $t$ is proposed to change customer $k$ for customer $i$ the new revenue will be:

$$Revenue'(t,i) = Revenue(t,i) + c \tag{2}$$

In order to accept a reassignment, two different situations should be considered:

a) If $d_{tkD} < d_{tiD}$: The total new distance is higher. We assume that an economically rational taxi driver will accept this new assignment if she earns the same as before plus the normal fare for the extra distance, that is if:

$$Revenue'(t,i) = Revenue(t,k) + (d_{tiD} - d_{tkD}) \cdot (fare - cost) \tag{3}$$

and thus,

$$c = Revenue(t,k) - Revenue(t,i) + (d_{tiD} - d_{tkD}) \cdot (fare - cost) \tag{4}$$

b) If $d_{tkD} \geq d_{tiD}$: The total new distance is the same or lower. In this case a driver should accept the reassignment if she earns the same as before (still, she would have to spend less time for the same income):

$$Revenue'(t,i) = Revenue(t,k) \tag{5}$$

and thus

$$c = Revenue(t,k) - Revenue(t,i) \tag{6}$$

Note that, depending on distances, the compensation $c$ may be positive or negative (i.e., it is received or has to be paid by taxi drivers). Based on the explained compensation scheme, a taxi coordination service can be defined that mediates between potential customers and taxis, assigns (re-assigns) customers to taxis, and pays/receives the corresponding compensations. The service implements the algorithm shown in Algorithm 1.

---

**Algorithm 1:** Original Taxi Assignment Algorithm [10]

---
1:     **Input**: current assignment $A^{current}$, accumulated *mediatorRevenue*
2:     **Output**: assignment $A^{new}$
3:     $A^o = A^{current} \cup \text{NTNR}(T^A, C^U)$
4:     $A$' = Calculate optimal assignment from $(C^A \cup C^U)$ to $T^D$
5:     **for all** $<t,i> \in A$' $\setminus A^o$ **do**
6:       **if** $d_{tkD} \geq d_{tiD} \mid <t,k> \in A^o$ **then**
7:          $c = Revenue(t,k) - Revenue(t,i)$
8:       **else**
9:          $c = Revenue(t,k) - Revenue(t,i) + (d_{tiD} - d_{tkD}) \cdot (fare - cost)$
10:      **end if**
11:       *mediatorRevenue* $- = c$
12:    **end for**
13:    **if** *mediatorRevenue* $\geq 0$ **then**
14:       $A^{new} = A$'
15:    **else**
16:       $A^{new} = A^o$
17:    **end if**
18:    **return** $A^{new}$

---

First, new unassigned customers ($C^U$) are assigned to available taxis ($T^A$) using an NTNR method (line 3). Then, an optimal global assignment $A$' is calculated considering only assigned (not occupied, i.e. dispatched) taxis ($T^D$) and assigned ($C^A$) or unassigned ($C^U$) customers (line 4). For each new assignment, i.e. a customer is proposed to be assigned to a different taxi, calculate the corresponding compensation $c$ (lines 5-12). Finally, if the sum of all compensations is not negative, and all involved taxis agree to the reassignment, then implement the new assignment (line 13). Otherwise, continue with the current assignment (line 16).

In this algorithm, a new assignment is only implemented, if the overall benefit of the mediator service is not negative (i.e., taxis pay more compensations than they receive). Thus, the mediator service does not incur in additional cost. Instead, it may have some benefit. With respect to the optimization step (line 4), different strategies have been proposed that minimize either the global distances, maximize the economic benefit of the mediator, or combine both ideas.

In the evaluation experiments (with different generated taxi and customer distributions) the proposed reassignment mechanism shows better performance in all parametrizations than classical assignment strategies (e.g., FCFS and NTNR) with respect to overall cost reduction (and thus, an increment in the global income adding taxi and coordinator service revenues) and average customer waiting times. The improvements depend strongly on the demand.

## 3.2    High level description of system operation

The goal of this subsection is to extract key aspects of the operation of the taxi coordination system outlined above that may be subject to special legal considerations. We aim at a description of system operation that is easily understandable by legal experts, hiding away pure technical details. For this purpose, for each of the high-level concepts put forward in [9], we provide a textual description for the application sketched in the previous subsection. The result is shown in Table 1.

**Table 1.** High-level concepts of system operation and their description

| Concept | Description |
|---|---|
| Subdomain | Regulation of taxi operation |
| Affected entities | –   taxi services |
| | –   customers of taxi services |
| | –   taxi coordination services |
| Definitions | see Table 2 |
| Operational facts | see Table 3 |
| Assumptions | see Table 4 |
| Affected general issues | –   Responsibility of service provision |
| | –   Data protection |
| | –   Transparency and accountability |

Notice that the table makes reference to definitions, operational facts, and assumptions that are described in Table 2, Table 3 and Table 4, respectively. Even though formal languages with rigorous semantics and proof theory could have been used to specify the operation principles of the proposed system, for the purpose of legal analysis our structured natural language description is deemed more appropriate.

**Table 2.** Definitions.

| Term | Description |
|---|---|
| Taxi coordination services | A private or public service that coordinates the assignment of taxis to customers. It can be freely used by customers to request taxi services. Taxi drivers can freely join a coordination service, which implies that they can be assigned through the coordination service to available customers. |
| Taxi | Vehicle (with driver) that offers transportation services with specific regulations (and pricing schemes). |
| Taxi states | *Available* (the taxi in not carrying a customer nor has committed to carry a particular customer), *assigned* (the taxi has committed to carry a customer but is not carrying her yet), or *busy* (the taxi is carrying a customer). |
| Customer states | *Unassigned* (not yet assigned to any taxi), *assigned* (assigned to a taxi, but still waiting for the taxi to arrive), *served* (currently in a taxi towards her destination location). |

**Table 3.** Operational facts of the taxi reassignment system.

| Fact | Description |
|---|---|
| F-1 | The taxi coordination service just assigns customers to taxis. It is not itself a transportation company. New customer requests are assigned to the closest available taxi. |
| F-2 | A taxi assigned to a customer may be reassigned to any other customer who is either closer or further away from the taxi's current location. |
| F-3 | If a taxi is reassigned to a customer that is closer to its current position than the previous customer, the taxi driver has to pay the estimated amount of cost of the extra driving cost he is saving by moving towards the closer customer instead of the customer that he was originally assigned to. |
| F-4 | If a taxi is reassigned to a customer that is farther away to the taxi's current position than the previous customer, the taxi driver receives a compensation for the extra arrival distance that is equal to the fare per kilometres times the extra distance (in kilometres). |
| F-5 | When the coordination service proposes a reassignment to a taxi driver, the taxi driver can accept or reject the proposed reassignment. |
| F-6 | If all affected taxi drivers accept a reassignment, it is implemented. |
| F-7 | The coordination service makes benefit through the surplus taxis pay when they are reassigned to a closer customer. |
| F-8 | Taxi customers pay the regular fare to the driver who is providing the taxi service. |
| F-9 | For each trip, a taxi driver receives the fare paid by the customer. In addition, the driver may receive or may have to pay a compensation to the coordination service (as specified in facts 3 and 4). |

**Table 4.** Assumptions

| ID | Description |
|---|---|
| A-1 | The fare scheme of the taxi service is based on a fixed rate that customers have to pay and that depends on the distance (and/or time) of the taxi ride from the initial position of the customer to its destination location. Additionally, there may be a fixed amount for starting a taxi trip. |
| A-2 | The cost for a taxi to move to the customers initial location is covered by the taxi driver. |
| A-3 | Customers cannot select the particular taxi (and driver) they want to use. |

## 4    Legal Analysis

In this section, we first summarise the current regulation of taxi transportation services in the city of Madrid, Spain. Then, we analyse as to how far the proposal described in section 3 complies with the current regulation and discuss several issues that need to be considered by coordination services like the one presented in section 3.

### 4.1 Current Regulation of Taxi Services

In Spain, like in many other countries, the legal-administrative regime of taxi services is usually regulated at the municipality level [34, 35]. It states that "the taxi does not properly constitute a public transport service, but an activity of general interest whose promotion, within the framework of the autonomous legislation, corresponds to the municipality" [36]. This transport activity is intervened and regulated by the Public Administration, but there is no public ownership of it. In the case of Madrid, which we use as a reference, the Ordinance of the City Council regulating the taxi sector continues to define this means of transport as a public service. However, it should be conceived as an "improper" public service, i.e., an activity of general interest that an individual provides by administrative authorization [37].

The main question, in the legal scope, is whether the proposal contained in section 3 conforms to Law, in particular to regulations of the Autonomous Community of Madrid and the City Council of Madrid. This proposal would fit in section no. 66 of the Decree of the Community of Madrid, No. 74/2005, of July 28, which approves the Regulation of Urban Public Transport Services in Passenger Vehicles where it is urged to promote the research, development, and implementation of new communication technologies in the taxi sector. Also, this proposal favours the environmental, energy efficiency and efficiency objectives of the sector (such as minimizing the waiting time) principles set by taxi regulations.

A key aspect of the proposal is the legality of a taxi mediation service (F-1). In this regard, the current regulations facilitate the establishment of intermediation companies through computer applications [38]. An intermediation company could perfectly use this system of reassignment as long as it complies with the official rates, established by the City Council of Madrid, that is, no surcharges can be established that increase the price to be paid by customers (F-8). Mediation services can establish their own internal operation (e.g. taxi assignment method) as long as they fulfil this requirement. Therefore, there is no direct incompatibility between the proposed functioning of the taxi coordination service (operational facts F-2, F-3, F-4, F-5, and F-6) and the current taxi regulations. In fact, intermediation companies could even establish that taxi drivers have to accept any proposed (re)assignments as a condition to participate in the platform.

Mediation companies can be public or private entities and thus generate benefit (F-7). Furthermore, they can establish discounts for customers for using their platform. They may include price reductions or bonuses for customers without them being considered as a violation of free and fair competition as indicated in judgment no. 956 of the Provincial Court of Barcelona, of 21 May 2019. The discounts or bonuses for customers do not imply, currently, a position of dominance in the taxi sector, and mediation companies subsidizing customers is not considered anticompetitive behaviour.

The taxi fares are usually regulated by the municipalities and typically are based on a basic fee plus a rate that depends on distances and/or time travelled, among other factors. Thus, assumption A-1 is compatible.

There are two options for customers to get a taxi: hailing a taxi on the street where the customer is located (or a taxi stop) or booking one to pick up the customer at a specific location. In the latter case, the cost of moving to the pick-up location can be either covered by the taxi driver or by the customer. Thus the option assumed by the proposed system (A-2) is feasible. In particular, the Ordinance no. 8546/2407 of the City Council of Madrid, from December 17th, 2019, establishes a maximum amount that can be charged in case a taxi has been booked in advance. It also establishes that, since January 1st, 2020, pre-contracting with a maximum closed price is allowed: according to the instructions approved by the City Council of Madrid, the client is favoured by knowing already the maximum amount to be paid. Since the reassignment decision is based on the fare the customer pays (F-8 and A-1) and cost, our method is also compatible with establishing a closed price agreement with the customer. In this case, the mediation service would still be interested in (re)assigning the closest taxi to each customer, which is the same base of our current approach.

In relation to customers' rights, the Ordinance of the City Council (article 52) establishes that customers can request specific types of taxis (regular, accessible, kids, …). The mediation service must comply with such vehicle requirement in any reassignment. However, customers do not have the right to choose a particular driver (A-3) as long as the vehicle fulfils their requirements.

In the European legal field, it is necessary to consider the Regulation (EU) 2019/1150 of the European Parliament and of the Council of 20 June 2019 on promoting fairness and transparency for business users of online intermediation services. The taxi assignment approach should promote the fairness and transparency principles (article 1) among professional users of the application, which is a benefit of consumers. In this regard, the rights of consumers and users, which are included in the Royal Legislative Decree 1/2007, of 16 November, approving the consolidated text of the General Consumer and User Protection Act and other complementary laws, must also be considered.

According to the European regulation, the principle of transparency is crucial to promoting sustainable business relationships and to preventing unfair behaviour to the detriment of customers. The regulation seeks to ensure adequate transparency and the possibility of effective demands against online platform services, and build trust between other similar companies and consumers, creating a fair, predictable, sustainable, and trusted online business environment. The entrepreneurial freedom (art. 38 Spanish Constitution) should favour, as provided in the European regulation, a competitive, fair, and transparent online ecosystem where companies behave responsibly. This is essential for consumer welfare. However, the application of the contractual law between suppliers and professionals who offer services to consumers is important, so the contractual relationship exists if both concerned parties express their intention to be bound in an unequivocal manner on a durable medium, even without the existence of an explicit written agreement. The general conditions must be easily accessible to users, be written in plain and intelligible language, easily available to users in the pre-contractual and contractual stage of the online intermediation relationship with the provider. In their terms and conditions, a description of any differentiated treatment which they give, or might give, in relation to services offered to consumers must be indicated to avoid unequal treatment. That description shall refer to the main economic, commercial, or legal considerations for such differentiated treatment.

### 4.2 Legal Conclusions and Requirements for Novel Taxi Services

In general, according to the description in section 4.1, there are no strong legal limitations that would make the application of the proposed taxi reassignment approach or the creation of the described taxi service coordinator infeasible. Still, from the legal compliance analysis, there are some particular issues that affect the proposed coordination service and that are worth discussing.

First, regarding the **types of taxis** customers can select, as mentioned in section 4.1, the mediation service has to assure that the customer is served by the requested type of vehicle. However, the proposed assignment method does not deal with different vehicle types. This aspect can be included in steps 1 and 2 of the algorithm, by considering only compatible vehicles for each particular customer request, i.e. incompatible taxis are considered unavailable (or busy) for that specific user. Note that different types of vehicles are not necessarily mutually incompatible (e.g. an "Accessible taxi" can also act as a regular one).

Second, the regulation states that it has to be clear **who has provided a certain service**. That is, the provider of a service must be identifiable in order to establish liabilities due to possible service failure. In this sense, means for registering the assignments of taxis to customers have to be integrated in the deployed platform. However, it is out of the scope of the assignment algorithm.

Third, the taxi driver who participates in the platform does it with all its consequences. In particular, it is possible to oblige her to **accept any assignment / reassignment.** The taxi driver's freedom is essentially to decide whether or not to use the computer application. So, the proposed coordination service could directly impose any identified (beneficial) reassignment, what would clearly simplify the approach assuring its full efficiency increase. Even in this case, from an economically rational perspective, taxi drivers should be willing to accept any reassignment (with its corresponding compensations), as it has been discussed in section 3. Thus, taxi drivers will have a clear incentive to participate in the coordination platform.

In addition, the proposed application reduces the overall distance that needs to be driven by the taxi fleet. This means that the use of our platform, by means of the dynamic reassignment of customers, generates an economic surplus which can be used in several ways. Usually, we expect it to be shared in some way between the platform owners and the taxi drivers. One particularly compelling option would be to subsidize taxi drivers by lowering (or even abolishing) the commission that each of them must pay for the use of the computer application (currently, in Madrid, up to 12% in existing platforms such as Uber).

Finally, the local administration may want to encourage the adhesion of taxi drivers to the application to promote infrastructures of this type in the production apparatus according to the Smart Cities model. The City Councils can carry out promotion mechanisms by establishing special rates to finance the different mediation platforms. The local administration must act with equity and avoid excessive damage to taxi drivers against private hire drivers (e.g. Uber and Cabify), for example in unfair competition cases. However, notice that it is obviously acceptable to create economic incentives in the form of fiscal and/or financial aids for vehicles that do not contaminate, and also for platforms that reduce the overall amount of distance travelled and thus sustain urban mobility while reducing traffic jams and contamination.

## 5 Adapted Assignment Method After Legal Analysis

In this section we address the conclusions obtained from the legal analysis carried out in the previous section. We propose modifications to the initial assignment method when necessary, or argue why no change is actually needed. Then we present some experimental results to analyse the performance of the resulting approach.

### 5.1 Method Modification

In this section we analyse in detail the potential method modifications identified by the legal analysis. We begin by describing a way of dealing with different types of taxis and the modification of the original algorithm. Then, we discuss the implications on the algorithm in case of allowing customers pre-contracting taxi services with maximum closed prices. Next, the option of charging to the customer the cost of reaching the pick-up location is analysed. Finally, we identify the modifications needed if external funding (e.g. public subsidy) is available.

#### 5.1.1 Types of taxis

In their requests, customers can specify the type of taxis they desire. By "type of taxi" we mean a taxi that complies with certain requirements, such as being suitable for passengers with reduced mobility (Eurotaxi), offering a minimum number of seats, being female-friendly[1], etc.

In order to adapt our proposal to include this aspect we need to specify taxi service requests from customers, taxi characteristics (mostly vehicle features) and a process to check compliance between both.

Instead of a flat set of independent taxi characteristics, we opt for giving some structure. In particular, we assume a taxonomy representation of taxi characteristics $TC = (Ch, \subseteq)$, where $Ch$ is a set of characteristics and $\subseteq$ is a partial order subsumption relation to allow some basic implicit reasoning over specifications. If $x \subseteq y$ (for clarity we use infix notation) then $x$ is less restrictive than $y$ ($x$ is subsumed by $y$) and thus, a taxi compliant with $y$ is also compliant with $x$. For example, 7-seater taxis are able to provide 5-seater services (as long as they comply with the rest of constraints), which is represented in our taxonomy as *5-seater $\subseteq$ 7-seater*.

Different taxi systems may have a different set of characteristics that need to be modelled, which leads to particular instances of *TC*.

***Customer request specification***. A customer ($i$) request for a taxi service includes a set of requirements $r(i) = \{r_1, r_2, ..., r_n\}$, $r_i \in Ch$, all of which must be satisfied by the dispatched taxi. If $r(i) = \phi$ all taxis are potential candidates to carry out the requested ("normal") service.

---

[1] Taxis driven by women

***Taxi type specification***. Similarly, each operating taxi $t$ includes a specification of the service characteristics $t$ it is are able to fulfil, $\tau(t) = \{\tau_1, \tau_2, ..., \tau_n\}$, $\tau_i \in Ch$.

***Taxi-Customer compatibility.*** In order to know if a given taxi $t$ is able to provide a service requested by a customer $i$ (which we call *compatible(t,i)*), it must be checked that all requirements in the customer's request are included in the taxi specifications, either explicitly or implicitly (derived by subsumption transitivity).

Thus, we define *compatible(t,i)* as:

$$compatible(t,i) = \mathrm{r}(\mathrm{i}) \subseteq \tau^*(\mathrm{t}) \tag{7}$$

where $\tau^*(t)$ is the transitive closure of $\tau(t)$ defined as:

$$\tau^*(t) = \bigcup_{k=1}^{\infty} \tau^k(\mathrm{t}) \tag{8}$$

$$\tau^1(\mathrm{t}) = \tau(\mathrm{t}) \tag{9}$$

$$\tau^{n+1}(\mathrm{t}) = \{\mathrm{x} \mid \mathrm{y} \in \tau^n(\mathrm{t}) \wedge \mathrm{x} \subseteq \mathrm{y}\} \tag{10}$$

For example, assume the next taxonomy of taxi requirements:
$TC = (Ch, \subseteq)$, Ch = {Eurotaxi, 9-seater, 7-seater, 5-seater, female-friendly},
$\subseteq$ = {(5-seater,7-seater), (7-seater, 9-seater)}
$\tau(t)$ = {Eurotaxi, 7-seater}
$\tau^*(t)$ = {Eurotaxi, 7-seater, 5-seater}

***Taxi assignment and reassignment operations***

In our original algorithm (Algorithm 1) new customers are firstly assigned to available taxis using an NTNR method, then the resulting global assignment is optimized proposing reassignments among the dispatched taxis (on the way to pick up customers). However, if we consider taxi types, a straightforward application of NTNR may allocate a more equipped taxi (e.g. a Eurotaxi) to a user without restrictions while leaving a customer with special needs unserved because of the scarcity of that type of taxi.

We propose a variation of that approach. The basic idea is to process the unassigned customers ($C^U$) according to the "complexity" of their requests' requirements.

For this, we assume that, given a taxi characteristic taxonomy $TC = (Ch, \subseteq)$, there is a function $p$ that returns the *priority* of a concept from $Ch$, i.e. $p\colon Ch \rightarrow Number$. This function must fulfil that $\forall\, x, y$ if $x \subseteq y \rightarrow p(x) \leq p(y)$.

Algorithm 2 shows the new assignment algorithm. The requirements of the customers are included in a list that is ordered by their "complexity" (i.e. priority), without duplicates (lines 4-7). Then, in descendent order priority, each request "type" is analysed (line 8). The set of customers $C$ who requested that type (line 9) and the set of taxis $T$ that can provide that service (line 10) are assigned using NTNR (line 11). $C$ and $T$ are updated removing the assigned elements (lines 12-13).

| **Algorithm 2:** Taxi assignment |
|---|
| 1:    **Input**: set of available taxis $T^A$, set of unassigned customers $C^U$ |
| 2:    **Output**: assignment $A$ |
| 3:    $A = \phi$ |
| 4:    ChList = [ ]   // ordered list of requests requirements |
| 5:    **for all** $i \in C^U$ **do** |
| 6:        ChList = insertOrd(ChList, $r(i)$) |
| 7:    **end for** |
| 8:    **for all** $ch \in ChList$ **do** |
| 9:        $C = \{ i \mid ch = r(i) \}$  // customers who requested $ch$ |
| 10:       $T = \{ t \mid ch \subseteq \tau^*(t) \}$  // taxis capable of providing $ch$ |
| 11:       $A = A \cup \text{NTNR}(T, C)$ |
| 12:       $C = C \setminus \{i \mid <i,t> \in A\}$ |
| 13:       $T = T \setminus \{t \mid <i,t> \in A\}$ |
| 12:    **end for** |
| 13:    **return** $A$ |

To insert taxi requests requirements in a sorted list (line 6) we define a total order function $\rho$ that, given a pair of requests ($r(i)$, $r(j)$), prefers the one with the highest priority characteristic. If this is of equal priority, then the second highest characteristic is compared, and so on. This can be implemented by sorting both sets by priority, then compare one by one until one is greater than the other or no elements are left.

For example, consider the following priority,

    $p(Eurotaxi) > p(female\text{-}friendly) > p(9\text{-}seater) > p(7\text{-}seater) > p(5\text{-}seater)$,

as well as the following unassigned customer requests,

    $r(c_1) = \{\}$, $r(c_2) = \{Eurotaxi\}$, $r(c_3) = \{Eurotaxi, female\text{-}friendly\}$, $r(c_4) = \{9\text{-}seater\}$.

Then, the ordered list of requests would be [$r(c_3)$, $r(c_2)$, $r(c_4)$, $r(c_1)$].

The algorithm presented in Algorithm 2 is used as the initial assignment in line 3 of algorithm 1.

### 5.1.2    Precontracting trips

As stated in section 4, it may be allowed to pre-contract a taxi service with a maximum closed price. In this case, we have to deal with different pricing schemes for taxi $t$ attending client $k$:

    a)   Predefined price: where $Revenue(t,k) = price - cost \cdot d_{tkD}$. That is, in this case the taxi still would have to pay the cost of the movements.

    b)   The scheme defined in section 3.1 with:

    $Revenue(t,k) = fcost + fare \cdot d_{kD} - cost \cdot d_{tkD}$           (11)

The compensation scheme for assigning taxi $t$ from client $k$ to client $i$ would not change, regardless of how *Revenue(t,k)* and *Revenue(t,i)* are defined, since it is based on reducing taxi costs. Thus, taxi trips with a fixed price can be reassigned in the same way from one taxi to another, if this would reduce the overall distance. In such a case, taxi drivers would be interested in the reassignments because they do not lose. Therefore, both pricing schemes could coexist in the same system.

### 5.1.3    Cost of empty trip to pick up a customer

In the pricing scheme used in section 3.1 it is assumed that the cost of moving to a customer's location is assumed by the taxi driver. Nevertheless, the current taxi regulation allows establishing a pricing scheme in which the customer pays for the taxi movement to the pick-up location. In this case, the revenue of a trip for taxi $t$ attending customer $k$ is calculated as follows:

$$Revenue(t,k) = fcost + (fare - cost) \cdot d_{kD} \tag{12}$$

However, the compensation scheme does not change.

If $d_{kD} < d_{iD}$:
$$c = Revenue(t,k) - Revenue(t,i) + (d_{iD} - d_{kD}) \cdot (fare - cost) \tag{13}$$
Otherwise ($d_{kD} \geq d_{iD}$):
$$c = Revenue(t,k) - Revenue(t,i) \tag{14}$$

In particular, note that the additional distance needs to be taken into account in the compensation since it represents the extra time spent by the taxi driver.

The rationale behind the approach presented in section 3.1 is that taxi-customer reassignments can reduce taxi trip distances, thus reducing costs. Those savings are essentially invested in compensating taxi drivers to accept "worse" assignments (extra cost and time). If the cost of moving to pick-up locations is defrayed by the customers, then driving savings mostly affect the amounts customers pay (still, reducing driving times is beneficial for taxi drivers). The consequence is that, unlike the pricing scheme assumed in the rest of this work, compensations $c$ will never be negative (i.e. the taxi pays some cost savings to the mediator). Therefore, no reassignments will be implemented in practice unless the mediator is allowed to have a negative budget balance or some of the policies described in section 5.1.4 is applied.

### 5.1.4    Compensation policy

Our proposal revolves around proposing assignment improvements that reduce the overall distance driven by taxis, resulting in lower users' waiting times, driving times, less pollution, etc. We incentivize taxis in such a way that they are willing to accept the reassignment proposal (assuming economically rational behaviour). The compensation budget is obtained from taxi cost savings paid to the mediator, so the latter only proposes improvements that keep this budget positive.

If public administrations would support mediation platforms with some level of funding, this could be used to increase the budget available for compensations, thus making feasible some "expensive" reassignment that would be discarded otherwise. The implementation of this option can simply be added to Algorithm 2 by initiating *mediatorRevenue* with some initial amount.

As we pointed out in section 4.2, it is perfectly legal that a mediation platform imposes to its participating taxis the acceptance of any assignment or reassignment proposal without any compensation. In this situation, the taxi drivers' choice is simply whether or not they want to join the platform. The implementation of this approach is straightforward as all management of compensations would just be removed from the Algorithm 2.

### 5.2 Experimental Evaluation

We carried out several experiments in order to evaluate how the changes introduced in the assignment algorithm affect its effectiveness. In these experiments we used the same simulation environment and setup as described in [10] and which we summarize in the following. A more detailed analysis of additional aspects of our initial assignment strategy can be found in [10].

We created synthetic simulations as follows. We simulated a taxi fleet of 1000 taxis in an area of about 9×9 km (roughly spanning the city of Madrid, Spain) during 5 hours of operation with randomly generated customers. Movements of taxis are simplified to straight-line movements with a constant speed of 17 km/h. We have approximated this value by dividing the average speed of cars in the city centre of Madrid in the last years (about 24 km/h) by 1.41 (the length of the straight line between opposite corners of a one-unit square, i.e., square root of 2). This adaptation is intended to better approximate the durations of trips in a straight-line simulation to trip durations in a real road network. Taxis are initially distributed randomly in the area with a uniform distribution and do not cruise (e.g., when idle they wait until a new service request). A fixed number of customers is generated (randomly) in each 15-minute interval. In order to analyse different supply/demand ratios, we vary this number from 250 to 1000 per 15 minutes, in steps of 125. Within the 9×9 km square we specified a central area. The origin and destination positions of customers are generated in the following way: each trip goes either from the outskirts to the centre, or vice versa. The points themselves are generated using a normal distribution (either in the centre or in the outside of the area). We also carried out experiments with a uniform distribution, where both, destinations and origins are generated using a uniform probability distribution over the whole area. The results are very similar and are thus omitted here.

In the experiments, we compare the results for the *MinDist/MaxRev* approach from [10] with the baseline methods *FCFS* and *NTNR*. *MinDist/MaxRev* combines a maximization of mediator benefit with minimizing distance. It is the strategy that has shown the best performance in [10] in terms of average client waiting times and overall distance reduction.

During a simulation, the assignment of taxis to customers is accomplished every 5

seconds using the corresponding assignment strategy. Taxis move towards the customers and eventually carry the customers to their final destinations. Pick-up and drop-off times are set to 30 and 90 seconds, respectively. The used payment scheme has a fixed cost of 2.4 euros per trip and each kilometre a customer moves with the taxi is paid by 1.05 euros. We assume a cost of operation of a taxi of 0.2 euros per kilometre.

Each experiment is repeated 10 times with a different randomly generated customers and the presented results are averages over the 10 runs.

In the first experiments we tested the behaviour of the modified assignment approach, including different types of customer requests and different types of taxis. In particular, we specified three different special taxi requests:

- Female-friendly taxi (20% of all customers)
- Taxi with capacities for wheelchair users; called Eurotaxi in Spain (10% of users)
- Female-friendly and Eurotaxi (5% of customers)

The last type simulates the idea that some customers could request several special requirements at the same time. The remaining 65% of users just request any taxi. With respect to the taxis, we specified also 20% to provide female-friendly services, 10% to Eurotaxis, and 5% of the taxis with both, Eurotaxi and female-friendly. As described before, the 5% female-friendly Eurotaxis can provide services for any customer, even for customers that do not have special requirements.

Fig. 2 compares the average waiting times (in minutes) for the different types of customer requests. In the case of *FCFS* and *NTNR* the assignment is also done based on a preference order of the customers that are waiting at a certain point of time. That is, first, customers asking for both, female-friendly and Eurotaxi are assigned. Afterwards, customers requesting Eurotaxis and then, customer requesting female-friendly taxis are assigned. Finally, the rest of customers are assigned to available taxis.

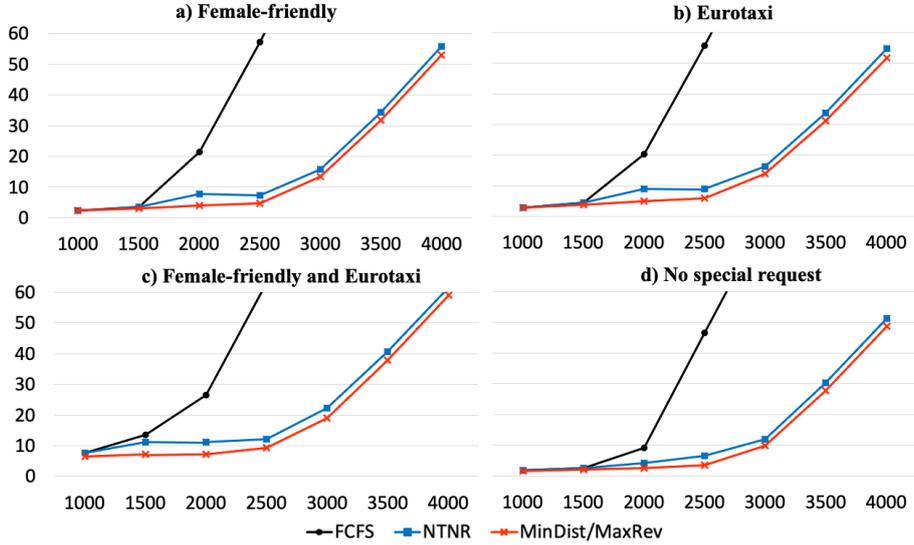

**Fig. 2.** Average waiting times (min) for different number of customers per hour and different types of customer requests.

As it can be seen, the *FCFS* approach performs very badly when the demand increases. In the case of a saturated system (e.g., the demand starts to exceed the supply, at about 2500 customers per hour) the waiting time for customers obviously increases. But *NTNR* and also *MinDist/MaxRev* are able to deal with this situation much better than *FCFS*. *MinDist/MaxRev* outperforms the baseline approaches in all cases, for special service requests and "normal" requests. The highest relative improvements regarding *NTNR* are obtained in situations where the supply and demand are roughly at the same level (with about 2000 /2500 customers per hour). For instance, in the case of 2000 customers per hour, *MinDist/MaxRev* improves the waiting time with regard to *NTNR* by 49.1% (from 7,8 to 4 min) for female-friendly requests, 45.7% (from 9.1 to 4.9 min) for Eurotaxis, 36.5% (from 11.2 to 7.1 min) for female-friendly Eurotaxis, and 39.8% (from 4.3 to 2.6 min) for "normal" requests. It is also noticeable that special requests will have higher waiting times, essentially because the number of taxis being able to attend such requests is lower.

As mentioned in section 4.2, public authorities may establish certain benefits of mediator services as the one analysed in this manuscript if, for example, they reduce the traffic and thus the pollution in an urban area. In the next set of experiments, we present the reduction of taxi movements applying the proposed reassignment strategy (*MinDist/MaxRev*) as compared to *FCFS* and *NTNR*. Here we do not differentiate different types of customer requests. In this sense, Fig. 3 presents the empty taxi movements in kilometres, i.e., the movements taxis have to do in order to reach the customers they attend.

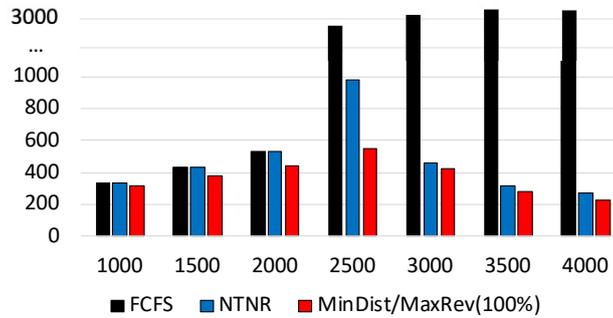

**Fig. 3**. Average distances of empty taxi movements (the values represent average values in kilometres and normalized to 1000 customers).

Fig. 3 confirms that the reassignment approach reduces the overall travel distances of taxis with respect to *NTNR* and for all different demand scenarios. Again, it can be observed that *FCFS* starts to perform badly when the demand starts to exceed the supply. It is interesting to observe that the highest reduction of *MinDist/MaxRev* over *NTNR* is again obtained when demand and supply are roughly the same. In the case of 2500 customers per hour, with *NTNR* 1000 taxis need about 978 km to reach 1000 customers. The proposed reassignment strategy, on the other hand, only needs 544 km (a reduction of 44%). Such a reduction clearly is beneficial for reducing the overall traffic in a city, and thus, public authorities may support the implementation of mediator services as the one proposed here or may establish some type of subvention for such services.

We also analysed how possible subventions could be used in the reassignment algorithm to obtain even better performances. However, no clear improvement can be obtained by increasing compensations or by setting an initial surplus of the mediator. Thus, we consider that possible subventions should either benefit customers (e.g., reducing the prices for customers) or could benefit taxi drivers. In both cases, the effect might be that more taxi drivers would be willing to participate in the proposed system. In practice, this would allow to obtain the reductions of taxi movements as presented in Fig. 3, where it is assumed that all taxis participate in the proposed system. In this regard, we also analysed the benefit taxi drivers would obtain if they did or they did not participate in the proposed reassignment system. Here, we divided the 1000 taxis in the experiment in two groups: 500 taxis that participate in the system and 500 that do not. In the first phase of the algorithm, the closest taxis are assigned to the waiting customers (regardless of whether or not a taxi participates in the reassignments). Afterwards, only the participating taxis will exchange their customers using the proposed algorithm. The results are presented in Fig. 4 (in this experiment we do not use different types of customer requests).

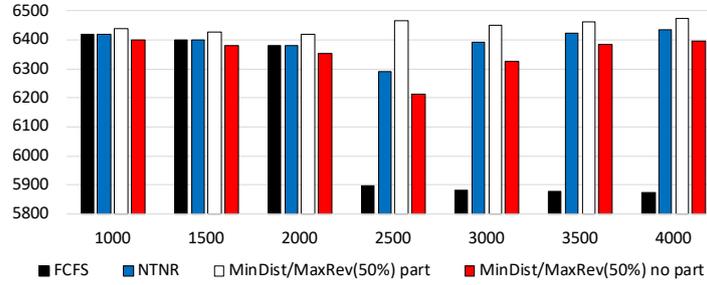

**Fig. 4.** Average taxi benefit obtained under different assignment strategies and participating or not participating in the reassignment system (the values represent monetary income in Euros normalized to 1000 customers)

As exposed in this experiment, even though the differences in the taxi incomes are rather low, the highest income is obtained by taxis that participate in the reassignment system. This is observed for all different demand values. Also, here, the difference between participating and non-participated taxis are highest when demand and supply are more or less at an equilibrium. It is likely that this rather reflects a real situation. As a conclusion of this experiment, from a rational point of view, a taxi driver should clearly participate in the proposed system, since her income would increase.

## 6    Conclusions

The area of Smart Cities is advancing at fast pace. The development of ICT, connectivity, Internet of Things, smart handheld devices, etc. has given rise to a huge number of developments that aim at providing efficient and sustainable solutions to daily life problems in big cities, like the management of traffic, for instance.

Many scientific solutions have been proposed without analysing their applicability in real life due to legal constraints, which can be different depending on the location in which the system is expected to be deployed. Closing the gap between law and a scientific solution may require adapting the solution, modifying the law, or a trade-off between both.

In this article, we have studied the legal implications of deploying collaborative taxi assignment strategies in a city. In particular, we set out from a proposal of a system that fosters collaboration among taxi drivers to exchange assigned customers, in order to reduce the overall movements of a taxi fleet. We analyse how that proposal fits with the current legislation in Spain, and in particular, in the city of Madrid.

Our main conclusion from the legal analysis is that the proposal presented in [10], in principle, could be deployed without a need for modifications in existing laws and ordinances. However, small adaptations of the algorithm are advisable in order to fulfil all legal requirements. We have modified the previous implementation to incorporate the recommendations identified by the legal analysis. Some of the options require very simple adaptations of the algorithm (including external funding for compensations) or no changes at all (setting a closed fare before starting a trip or charging the customer the cost of reaching the pick-up location). A more important modification was needed to incorporate different taxi types that can serve customer requests with special need (e.g., taxis for disabled people). We proposed a way of dealing with such cases as well as a modification of the original algorithm that keeps compliance of taxi-customer pairs in assignments and reassignments.

We carried out several experiments to evaluate the new implementation. We compared the well-known classical *FCFS* and *NTNR* approaches with our proposal (the new implementation of the best configuration observed in [10]). An analysis of the average waiting times of customers showed that, specially under high demand, our method outperforms *NTNR* in about 40% for every type of request, while *FCFS* performs very badly. We also observed that our method reduces the overall travel distances in comparison to *NTNR* (reaching a 44% improvement). Again, *FCFS* obtains very bad results with high demand. We tested the effect of injecting some funding into the mediation system to increase compensations, concluding that no relevant improvement can be obtained. Thus, our suggestion is that external (e.g. Municipality) subvention should be used to reduce prices for customers or benefit taxi drivers for participating in the system. Our last experiment confirmed that taxis participating in the reassignment system obtain higher incomes that those who decide not to join it.

Just as *NTNR*, the proposed method has a specific behaviour in case of service saturation (e.g., the number of customers exceeds the number of taxis). In such cases, it may happen that a customer waits for a long time to get a taxi. In the proposed algorithm, it is possible that a customer is unassigned after being assigned a taxi. If the customer is aware of that process the quality of service perception may be low. This can be easily dealt with by means of ad hoc solutions, like giving preference to customers who have been waiting for more than some predefined time limit.

Regarding future lines of research, we think it could be interesting to analyse situations where several service mediation entities coexist and compete. In such situations it might be of interest to foster cooperation among such entities, for example, through some kind of dynamic subcontracting. This might be useful in situations of scarcity of taxis or in order to reduce taxi movements globally. We think that multiagent systems and agreement technologies are good starting points for coordinating different mediation services, and modern multiagent frameworks such as SPADE [39] are promising candidates for the implementation of such a distributed platform.

Another interesting point is to incorporate the possibility of ride sharing into the proposed mediator service. Here, several users might share a taxi trip or taxis may pick up additional users during a service. From the legal perspective, ride sharing is allowed by section 37 of the Decree of the Community of Madrid, No. 74/2005, of July 28, which approves the Regulation of Urban Public Transport Services in Passenger Vehicles as amended by Decree 35/2019. This possibility would mean lowering the price to be paid by users, a better efficiency of the taxi sector, and reduction of traffic congestion and air pollution.

**Acknowledgments.** This work has been partially supported by the Spanish Ministry of Science, Innovation and Universities, co-funded by EU FEDER Funds, through project grant InEDGEMobility RTI2018-095390-B-C33 (MCIU/AEI/FEDER, UE).

## References

1. Botsman, R.: Defining the Sharing Economy: What is Collaborative consumption-and What isn't?, fastcoexist.com, available at http://www.fastcoexist.com/3046119/defining-the-sharing-economywhat-is-collaborative-consumption-and-what-isnt (2015).
2. Ossowski S. (ed) (2013) Agreement Technologies. Law, Governance and Technology Series (LGTS) no.8, Springer, Dordrecht.
3. Juan A. Rodríguez-Aguilar J.A, Sierra C., Arcos J.Ll,; López-Sánchez M., Rodríguez I. (2015) Towards next generation coordination infrastructures, Knowledge Engineering Review 30: 435-453
4. Fornara N, Lopes Cardoso H, Noriega P, Oliveira E, Tampitsikas C (2013) Modelling Agent Institutions. In: [2], pp 277-308.
5. Davidsson P., Gustafsson M., Holmgren J., Jacobsson A., Persson J (2013) Agreement Technologies for Supporting the Planning and Execution of Transports. In: [2], pp 533-547.
6. Billhardt H., Fernández A., Lujak M., Ossowski S. (2018) Agreement Technologies for Co-ordination in Smart Cities. Applied Science 8 (5)
7. Dötterl J.; Bruns B., Dunkel J., Ossowski, S. (2020): Evaluating Crowdshipping Systems with Agent-Based Simulation. In: Multi-Agent Systems and Agreement Technologies (Bassiliadis et al, eds.) . LNCS 12520, Springer, pp. 396-411
8. Casanovas P (2013) Agreement and Relational Justice: A Perspective from Philosophy and Sociology of Law. In: [2], pp 17-41.
9. Santos, J. A., Fernández, A., Moreno-Rebato, M., Billhardt, H., Rodríguez-García J. A., Ossowski, S. Legal and ethical implications of applications based on agreement technologies: the case of auction-based road intersections. Artificial Intelligence and Law 28, 385–414 (2020). https://doi.org/10.1007/s10506-019-09259-8
10. Billhardt, H., Fernández, A., Ossowski, S., Palanca, J., Bajo, J.: Taxi Dispatching Strategies with Compensations. Expert Systems with Applications 122, 173-182 (2019).
11. Egbelu, P.J., Tanchoco, J.M.A. (1984): Characterization of Automatic Guided Vehicle Dispatching Rules. International Journal of Production Research 22 (3), 359–374
12. Maciejewski M., Bischoff J. (2015). Large-scale Microscopic Simulation of Taxi Services. Procedia Computer Science 52, 358–364
13. Wittmann M, Neuner L, Lienkamp M (2020). A Predictive Fleet Management Strategy for On-Demand Mobility Services: A Case Study in Munich. Electronics 9(6):1021
14. Lee, D., Wang, H., Cheu, R., & Teo, S. (2004). Taxi dispatch system based on current demands and real-time traffic conditions. Transportation Research Record: Journal of the Transportation Research Board, 1882 (1), 193–200.
15. Maciejewski, M., Bischoff, J., & Nagel, K. (2016). An Assignment-Based Approach to Efficient Real-Time City-Scale Taxi Dispatching. IEEE Intelligent Systems, 31 (1), 68–77.
16. Zhu, C., & Prabhakar, B. (2017). Reducing Inefficiencies in Taxi Systems. In Proc. of the 56th IEEE Conference on Decision and Control (CDC). (pp. 6301-6306)
17. Zhan X., Qian X., Ukkusuri, S.V. (2016) A Graph-Based Approach to Measuring the Efficiency of an Urban Taxi Service System. IEEE Transactions on Intelligent Transportation Systems 17(9), 2479–2489
18. Vazifeh M.M., Santi, P., Resta, G., Strogatz, S.H., Ratti, C. (2018). Addressing the minimum fleet problem in on-demand urban mobility. Nature 557, 534–538
19. Dai, G., Huang, J., Wambura, S. M., & Sun, H. A. (2017). Balanced Assignment Mechanism for Online Taxi Recommendation. In Proc. of the 18th IEEE International Conference on Mobile Data Management (MDM) (pp. 102–111).
20. Gao, G., Xiao, M., & Zhao, Z. (2016). Optimal Multi-taxi Dispatch for Mobile Taxi-Hailing Systems. In Proc. of the 45th International Conference on Parallel Processing (ICPP) (pp. 294-303).

21. Meghjani, M., & Marczuk, K. (2016). A hybrid approach to matching taxis and customers. In Proc. of the Region 10 Conference (TENCON), (pp. 167-169).
22. Moreira-Matias, L., Gama, J., Ferreira, M., Mendes-Moreira, J., & Damas, L. (2013). Predicting taxi-passenger demand using streaming data. IEEE Transactions on Intelligent Transportation Systems, 14 (3), 1393–1402.
23. Zhang, D., Sun, L., Li, B., Chen, C., Pan, G., Li, S., & Wu, Z. (2015). Understanding taxi service strategies from taxi gps traces. IEEE Transactions on Intelligent Transportation Systems, 16 (1), 123–135.
24. Miao, F., Han, S., Lin, S., Stankovic, J. A., Zhang, D., Munir, S., Huang, H., He, T., & Pappas, G. J. (2016). Taxi Dispatch With Real-Time Sensing Data in Metropolitan Areas: A Receding Horizon Control Approach. IEEE Transactions on Automation Science and Engineering, 13(2), 463-478.
25. Miao, F., Han, S., Lin, S., Wang, Q., Stankovic, J. A., Hendawi, A., Zhang, D., He, T., & Pappas, G. J. (2019). Data-Driven Robust Taxi Dispatch Under Demand Uncertainties. IEEE Transactions on Control Systems Technology vol. 27, no. 1, pp. 175-191
26. Ma, S., Zheng, Y., & Wolfson, O. (2013). T-share: A large-scale dynamic taxi ridesharing service. In Proc IEEE 29th International Conference on Data Engineering (ICDE) (pp. 410–421).
27. Li, J. P., Horng, G. J., Chen, Y. J., & Cheng, S. T. (2016). Using Non-cooperative Game Theory for Taxi-Sharing Recommendation Systems. Wireless Personal Communications, 88 (4), 761-786.
28. Tian, C., Huang, Y, Liu, Z., Bastani, F., & Jin, R. (2013). Noah: a dynamic ridesharing system. In Proceedings of the 2013 ACM SIGMOD International Conference on Management of Data (pp. 985–988).
29. Billhardt, H., Lujak, M., Sánchez-Brunete, V., Fernández, A., Ossowski, S. Dynamic coordination of ambulances for emergency medical assistance services. Knowledge-Based Systems 70, 268-280, 2015
30. Glaschenko, A., Ivaschenko, A., Rzevski, G., & Skobelev, P. (2009). Multi-Agent real time scheduling system for taxi companies. In Proc. Int. Conf. Autonomous. Agents and Multiagent Systems (AAMAS), (pp. 29-36).
31. Dignum, V. (2020). AI is multidisciplinary. AI Matters, 5(4), 18–21. doi:10.1145/3375637.3375644
32. Dignum, V. (2017). Responsible Autonomy. Proceedings of the Twenty-Sixth International Joint Conference on Artificial Intelligence. doi:10.24963/ijcai.2017/655
33. Boella G, Humphreys L, Muthuri R, Rossi P, van der Torre L (2014) A Critical Analysis of Legal Requirements Engineering from the Perspective of Legal Practice, 2014 IEEE 7th International Workshop on Requirements Engineering and Law (RELAW), Karlskrona, 2014, pp. 14-21.
34. Doménech Pascual, G., Soriano Arnanz, A., Taxi Regulation in Spain Under the Pressure of the Sharing Economy, Uber & Taxis Comparative Law Studies, (Rozen Noguellou and David Renders Editors), Bruylant, (2018) p. 357
35. Mutiarin, D., Nurmandi, A., Jovita, H., Fajar, M. and Lien, Y.-N. (2019), "How do government regulations and policies respond to the growing online-enabled transportation service (OETS) in Indonesia, the Philippines, and Taiwan?", Digital Policy, Regulation and Governance, Vol. 21 No. 4, pp. 419-437.
36. Tarrés Vives, M.: La regulación del taxi, Atelier (2006), pp. 38-43.
37. Entrena Cuesta, R.: El servicio del taxi, Revista de Administración Pública, 27, 29-62 (1958).
38. OECD, International Transport Forum: App-Based Ride and Taxi Services: Principles for Regulation; Corporate Partnership Board Report (2016).
39. Palanca, J., Terrasa, A., Julian, V., Carrascosa, C. (2020): SPADE 3 – Supporting the New Generation of Multi-Agent Systems. IEEE Access 8, pp. 182537-182549